\documentclass{article} 
\usepackage{collas2025_conference,times}
\usepackage{easyReview}

\usepackage{amsmath,amsfonts,bm}

\def\eqref#1{equation~\ref{#1}}

\def\1{\bm{1}}

\def\ra{{\textnormal{a}}}

\def\rx{{\textnormal{x}}}

\def\rva{{\mathbf{a}}}

\def\erva{{\textnormal{a}}}

\def\ervx{{\textnormal{x}}}

\def\rmA{{\mathbf{A}}}

\def\vmu{{\bm{\mu}}}
\def\vtheta{{\bm{\theta}}}
\def\va{{\bm{a}}}

\def\ve{{\bm{e}}}

\def\vx{{\bm{x}}}

\def\eva{{a}}

\def\mA{{\bm{A}}}

\def\mH{{\bm{H}}}
\def\mI{{\bm{I}}}
\def\mJ{{\bm{J}}}

\def\mX{{\bm{X}}}

\def\mSigma{{\bm{\Sigma}}}

\DeclareMathAlphabet{\mathsfit}{\encodingdefault}{\sfdefault}{m}{sl}
\SetMathAlphabet{\mathsfit}{bold}{\encodingdefault}{\sfdefault}{bx}{n}
\newcommand{\tens}[1]{\bm{\mathsfit{#1}}}
\def\tA{{\tens{A}}}

\def\tX{{\tens{X}}}

\def\gG{{\mathcal{G}}}

\def\sA{{\mathbb{A}}}
\def\sB{{\mathbb{B}}}

\def\sS{{\mathbb{S}}}

\def\emA{{A}}

\newcommand{\etens}[1]{\mathsfit{#1}}

\def\etA{{\etens{A}}}

\newcommand{\E}{\mathbb{E}}

\newcommand{\R}{\mathbb{R}}

\newcommand{\KL}{D_{\mathrm{KL}}}
\newcommand{\Var}{\mathrm{Var}}

\newcommand{\Cov}{\mathrm{Cov}}

\newcommand{\normltwo}{L^2}
\newcommand{\normlp}{L^p}

\newcommand{\parents}{Pa} 

\usepackage{hyperref}
\hypersetup{
    colorlinks=true,
    linkcolor=red,
    filecolor=magenta,
    urlcolor=blue,
    citecolor=purple,
    pdftitle={Overleaf Example},
    pdfpagemode=FullScreen,
    }

\newcommand{\method}{\tt{L-RAPM}}
\newcommand{\improv}{\delta}
\newcommand{\lineup}{\lambda}
\newcommand{\lineupset}{\Lambda}
\usepackage{xcolor} 

\title{Lineup Regularized Adjusted Plus-Minus ({\method}): Basketball Lineup Ratings with Informed Priors}

\author{Christos Petridis \\
Department of Computer and Information Sciences\\
Temple University\\
Philadelphia, United States \\
\texttt{christos.petridis@temple.edu} \\
\And 
Konstantinos Pelechrinis  \\
Department of Informatics and Networked Systems \\
University of Pittsburgh \\
Pittsburgh, United States \\
\texttt{kpele@pitt.edu}
}

\collasfinalcopy
\begin{document}

\maketitle

\begin{abstract}
Identifying combinations of players (that is, lineups) in basketball - and other sports - that perform well when they play together is one of the most important tasks in sports analytics. 
One of the main challenges associated with this task is the frequent substitutions that occur during a game, which results in highly sparse data. 
In particular, a National Basketball Association (NBA) team will use more than 600 lineups during a season, which translates to an average lineup having seen the court in approximately 25-30 possessions. 
Inevitably, any statistics that one collects for these lineups are going to be noisy, with low predictive value. 
Yet, there is no existing work (in the public at least) that addresses this problem. 
In this work, we propose a regression-based approach that controls for the opposition faced by each lineup, while it also utilizes information about the players making up the lineups. 
Our experiments show that {\method} provides improved predictive power than the currently used baseline, and this improvement increases as the sample size for the lineups gets smaller.
\end{abstract}

\section{Introduction}
\label{sec:intro}

Given an opponent's lineup $\lineup_O$ and a set $\lineupset$ of possible lineups that our team can play, by how many points do we expect each lineup $\lineup$ $\in \lineupset$ to outperform $\lineup_O $? 
This is the core question we are trying to answer in this study by developing $\method$, a regression-based model (described in the following sections in detail). 
Although this is one of the most important tasks for the coaches and analytics departments of teams, there is very little public work that allows someone to answer this question to a satisfactory degree. 
A lineup is typically evaluated through its offensive, defensive and net rating, that is, the number of points it scores per 100 possessions (offensive rating), the number of points allowed per 100 possessions (defensive rating), and the difference between these two (net rating). 
These ratings are the only publicly available metrics for evaluating lineups and they are usually taken at a face value when it comes to projecting future lineup performance.

\begin{figure}
    \centering
    \includegraphics[width=0.45\linewidth]{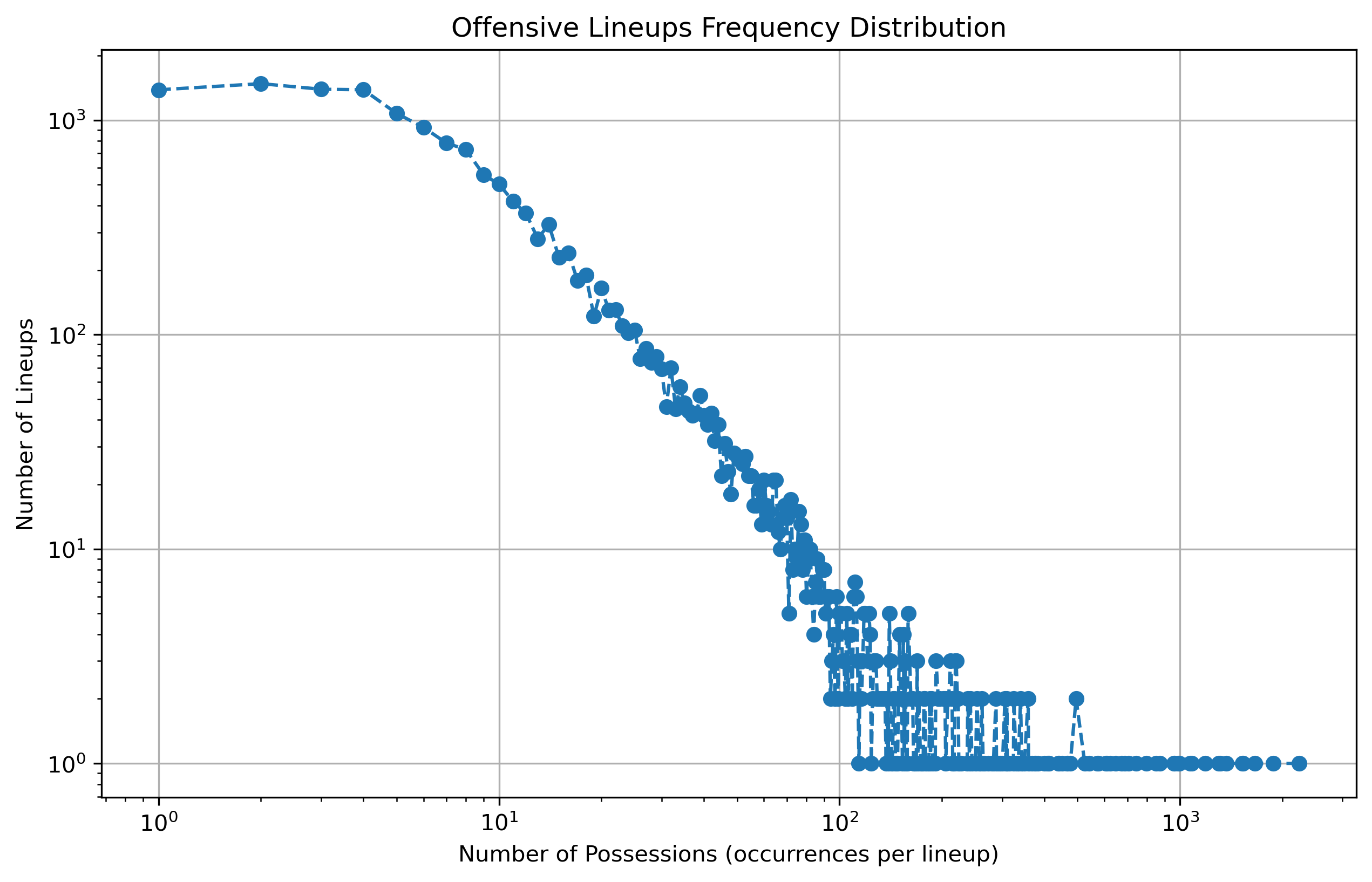}
    \includegraphics[width=0.45\linewidth]{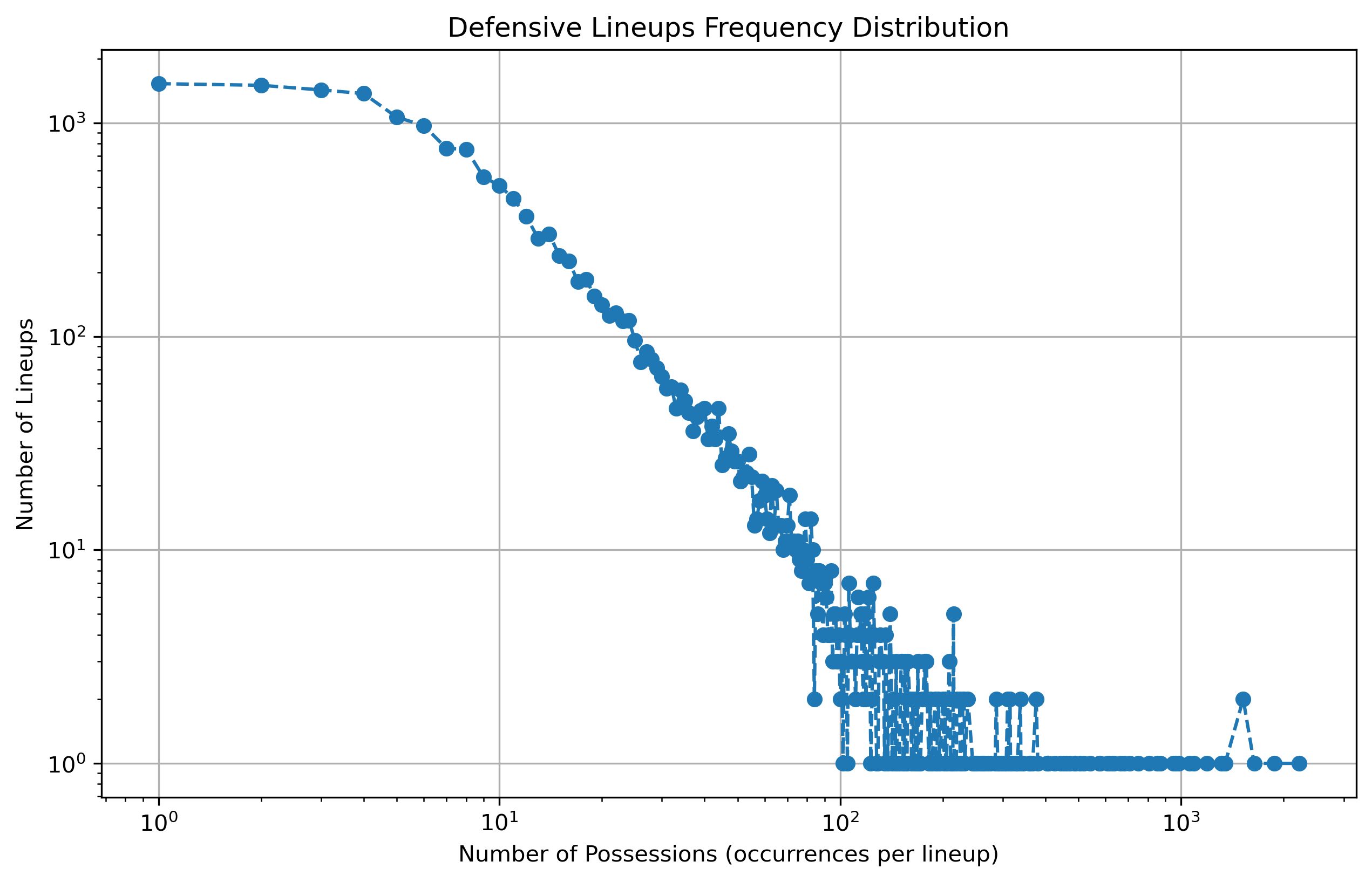}
    \caption{The majority of the lineups during a season appear on the court for a small number of possessions. The distribution (in log-log scale) for the number of offensive (left) and defensive (right) possessions per lineup is right skewed, with a small number of lineups playing for thousands of possessions, while the vast majority of them plays for less than 50 possessions. }
    \label{fig:poss_hist}
\end{figure}

While these raw (as we will call them for the rest of this work) ratings capture how a lineup has performed so far, they suffer from small - sometimes tiny - sample sizes, thus, limiting their predictive power. 
Using data from the 2023-24 NBA season, Figure \ref{fig:poss_hist} shows the distribution of the number of total possessions (both offense and defense) that each lineup played during the season. 
We use log-log scale since the distribution is right skewed, i.e., the vast majority of the lineups are on the court for very few possessions, while there is a small number of lineups that have played in thousands of possessions. 
More specifically, an average lineup plays approximately 17 possessions on offense and 17 on defense, with a standard deviation of 56 possessions. 
When dealing with this type of sample sizes, there is a lot of uncertainty associated with the raw ratings. 
For instance, let us consider lineup $\lineup_a$, that has played 10 offensive possessions and has scored 13 points. 
Its raw offensive rating is 113.0, i.e., $\lineup_a$ has scored at a rate of 113 points per 100 possessions. 
If, in one of the 10 possessions during which the lineup was on the court, a made three-point shot had instead resulted in a miss (which, on average, occurs 64\% of the time), the raw offensive rating of $\lineup_a$ would have been 100.0. 
The difference between offensive ratings of 113.0 and 100.0 is of the same magnitude as the difference between the offensive rating of the best and worst offensive teams in the league. 
Thus, it should be clear that using these raw lineup ratings to make projections for future performance is extremely noisy and not robust.

An additional limitation of using these raw ratings is that they are not taking into account the strength of opposition faced. 
For example, let us consider two lineups $\lineup_a$ and $\lineup_b$ that both have the same raw ratings. 
One might be tempted to say that these two lineups are of the same quality. 
However, what if you had some additional information with regards to who each lineup has faced on the court? 
What if $\lineup_a$ played all of its possessions against the best players of the opposing teams, while $\lineup_b$ mainly played during the end of games that were already decided, and the opponent had inserted their ``end-of-bench'' players?
This additional information would most probably lead you to consider $\lineup_a$ as a {\em better} lineup to $\lineup_b$. 

In this work, we address these challenges by developing $\method$, a regression-based model that controls for the opposition faced by a lineup, while also using a prior for each lineup to overcome the small sample size problems. 
These priors are informed from individual player ratings and they are incorporated in our model through regularization. 
Our experiments showcase the benefits $\method$ provides in terms of prediction, especially for lineups that have not played many possessions (which is the majority of the lineups as per Figure \ref{fig:poss_hist}). 
More specifically, for lineups that have played several hundred possessions, using their raw offensive and defensive ratings does almost as good of a job at predicting points per possession as {\method}. 
However, as we start dealing with lineups that have played very little time - e.g., less than 50 possessions, or even they have not played at all - the benefits from {\method} increase. 

The rest of this paper is organized as follows: 
Section \ref{sec:related} discusses relevant studies to our paper and further differntiates our work. 
Section \ref{sec:method} describes our method in detail, while Section \ref{sec:results} describes the data we used for our evaluations as well as our results. 
Finally, Section \ref{sec:conclusions} concludes our work, discussing possible directions for further improvements. 

\section{Related Studies}
\label{sec:related}

While there is a significant volume of research on evaluating players, and decomposing their impact from that of their teammates, there is little public work on the opposite direction, that is, given a lineup to project its performance. 
In terms of player ratings, the current gold standard includes a plethora of metrics that are based on adjusted plus-minus (APM) \citet{winston2022mathletics,adjpm}.  
A player's APM is calculated through a regression model, where each data point corresponds to a possession and the dependent variable is the points scored during the possession. 
The independent variables correspond to the players, each of whom is associated with two binary variables, one for offense ($x_{i,off}$) and one for defense ($x_{i,def}$). 
All independent variables are 0 except for the players in the lineups. 
For instance, if player $i$ is part of the offensive lineup for data point $j$, then $x_{j,i,off}=1$. 
Solving this regression allows us to divide credit among players for their team performance (points scored/allowed per possession), adjusting for who else was on the court with them. 
More specifically, the coefficient for variable $x_{i,off}$ corresponds to how many points per possession player $i$ contributes to his team's offense, while the coefficient for $x_{i,def}$ does the same for player's $i$ team's defense.  
One of the problems of the early versions of APM is that they were mainly descriptive metrics, capturing what has happened until that moment, and had little predictive power. 
To solve this problem \citet{sill2010improved} introduced the regularized version of the APM model (RAPM) that has more predictive power. 
While \citet{sill2010improved} shrank the coefficients to 0 through the regularization, other models (e.g., ESPN's RPM, 538's RAPTOR etc.) that have appeared since utilize a different prior value for each coefficient to shrink to based on a variety of other factors such as box score statistics \citet{winston2022mathletics,deshpande2016estimating}. 
In our study, we will make use of \citet{sill2010improved}'s RAPM for player ratings, since most of the metrics that use some type of box score prior are proprietary. 

When it comes to evaluating lineups the most prevalent approach - as mentioned in the previous section - is through the (raw) offensive, defensive, and net rating that are readily available from the league website \citet{nba-eff}. 
A basic adjustment has also been introduced \citet{winston2022mathletics} for the raw offensive, defensive and net ratings of a lineup, by subtracting the average player ratings of the opponents faced. 
For example, let us assume that $\lineup_a$ has a raw defensive rating of 113.2, and the average offensive rating of the players faced is +0.26 (per 100 possessions). 
This means that $\lineup_a$ has faced on average players that are better than average, and we can adjust the defensive rating of $\lineup_a$ to $113.2-(5*0.26) = 111.9$. 
While this is certainly an improvement over the use of the raw lineup ratings, it still suffers from small sample sizes and extreme outliers. 
A network embedding approach was introduced by \citet{pelechrinis2018linnet}, where the nodes of the network are the different lineups and there is a direct edge from $\lineup_a$ to $\lineup_b$ if $\lineup_b$ outperformed $\lineup_a$, weighted by the net rating. 
The authors then use {\tt node2vec} to obtain an embedding for the nodes/lineups that can later be used for downstream tasks, such as predicting matchups. 
This approach is also susceptible to small sample sizes, that can lead to noisy lineup embeddings. 
Furthermore, this model is not able to make predictions for previously unseen lineups, as these lineups have not been part of the training network. 

In the following section we will present in detail our proposed model, {\method}, and describe how each element addresses the challenges of small sample sizes and opponent strength. 

\section{Our model}
\label{sec:method}

$\method$ consists of 2 modeling components that we describe in the rest of this section: (a) individual player ratings, and, (b) lineup ratings. 

\subsection{Player ratings: Regularized Adjusted Plus Minus}
\label{sec:rapm}

In order to build the priors for our lineup ratings we need to first obtain ratings for individual players. 
For this purpose, we will use \citet{sill2010improved}'s RAPM model. 
In our study, we will use the season prior to the one for which we rate lineups to obtain the player ratings. 
For instance, if we are interested in obtaining lineup ratings for the season 2023-24, we obtain our player ratings using data from the 2022-23. 
There are several other options, with their own pros and cons, that we discuss in Section \ref{sec:conclusions}. 

To obtain RAPM we need the following information for each possession: (i) players on offense, (ii) players on defense, (iii) points scored. 
Every player $p$ is associated with 2 {\em dummy} variables in the model, $x_{p,off}$ and $x_{p,def}$. 
For each data point $i$, if player $p$ is part of the offense $x_{i,p,off}=1$, while if he is part of the defense $x_{i,p,def}=-1$. 
In any other case $x_{i,p,off}=0$ and $x_{i,p,def}=0$. 
With $\mathbf{b}$ being the linear regression coefficients that correspond to the player ratings, and $\mathbf{y}$ the dependent variable vector (i.e., $y_i$ is the points scored during possession $i$), we solve the following optimization problem: 

\[
\min_{\bm{\gamma}} \; \sum_{i=1}^{n} \left( y_i - (\gamma_0+\sum_{j=1}^{m} \gamma_{j,off}\cdot x_{i,j,off}+\sum_{j=1}^{m} \gamma_{j,def}\cdot x_{i,j,def})\right)^2 + \lambda_{off}\sum_{j=1}^{m} \gamma_{j,off}^2 + \lambda_{def}\sum_{j=1}^{m} \gamma_{j,def}^2 
\]
where $n$ is the number of data points and $m$ is the number of players. 
We use a different regularization constant for the offensive and defensive coefficients/ratings for the players to provide our model with more flexibility. 
We identify the appropriate values for $\lambda_{off}$ and $\lambda_{def}$ through a validation set. 
With this set up, $\gamma_0$ corresponds to the league average points per possession scored. 
$\gamma_{p,off}$ captures the points per possession above league average that player $p$ contributes when on offense. 
A positive value corresponds to a player better than average on offense. 
Similarly, $\gamma_{p,def}$ captures the points per possession below league average that player $p$ {\em saves} when on defense. 
A positive value corresponds to a player better than average on defense. 

Using data from the 2022-23 NBA season, we find that the values that minimize the validation error are $\lambda_{off} = 4000$ and $\lambda_{def}=6000$. 
The player ratings obtained at this step, will then be used to create the informed prior for the lineup regression model described below. 

\subsection{Regularized Regression for Lineup Matchups}
\label{sec:lineup_reg}

The second component of {\method} is a regression model for the points scored per possession, where the independent variables are the offensive and defensive lineup units (rather than the individual players). 
While this model adjusts for the opponents/lineups faced, the sparsity of the data limits the prediction power of the results. 
For this reason we will use regularization to help the model and its predictive ability. 
However, unlike with RAPM we are going to shrink the coefficients for the lineups not to 0, but to a different value, informed by the player ratings. 

{\bf Prior values} $\bm{\pi}$: Let us assume lineup $\lineup_a$ that consists of players $p_1, p_2,p_3,p_4,p_5$. 
If we did not know anything about the players in the lineup, we could start by making the assumption that this is a lineup that will perform at a league-average level. 
Nevertheless, we do have information about the players, and in particular we have their RAPM players ratings. 
Therefore, we can start with the belief that before seeing any data for $\lineup_a$, its offensive and defensive ratings would be:

\[
\pi_{\lineup_a,off} = league\_ppp + \gamma_{1,off} + \gamma_{2,off} + \gamma_{3,off} + \gamma_{4,off} + \gamma_{5,off} 
\]
\[
\pi_{\lineup_a,def} = league\_ppp + \gamma_{1,def} + \gamma_{2,def} + \gamma_{3,def} + \gamma_{4,def} + \gamma_{5,def} 
\]
where $league\_ppp$ is the league-average points scored per possession. 
Therefore, with $\bm{\beta}$ being the set of offensive and defensive lineup coefficients for our model, we have the following optimization problem: 

\[
\min_{\bm{\beta}} \; \sum_{i=1}^{n} \left( y_i - (\beta_0+\sum_{j=1}^{l} \beta_{j,off}\cdot x_{i,j,off}+\sum_{j=1}^{l} \beta_{j,def}\cdot x_{i,j,def})\right)^2 + \lambda\cdot\sum_{j=1}^{l} (\beta_{j,off}-\pi_{j,off})^2 + \lambda \cdot \sum_{j=1}^{l} (\beta_{j,def}-\pi_{j,def})^2 
\]
where $l$ is the number of different lineups, $n$ is the number of data points, and $x_{i,j}$ are regression variables indicating whether lineup $\lineup_j$ is on offense, defense or not involved in data point $i$. 
The coefficients $\bm{\beta}$ we will obtain from solving the above optimization problem are the final lineup ratings that we can use for predictions. 
These ratings have been adjusted for the opponents that each lineup has faced through the regression covariates (both on offense and defense), and they are also informed from individual player ratings through the shrinkage process of the regularization. 
In the following section, we will evaluate the predictive power of {\method} and compare it with the a baseline that uses the raw lineup ratings. 

\begin{figure}[ht]
    \centering
    \includegraphics[width=0.75\linewidth]{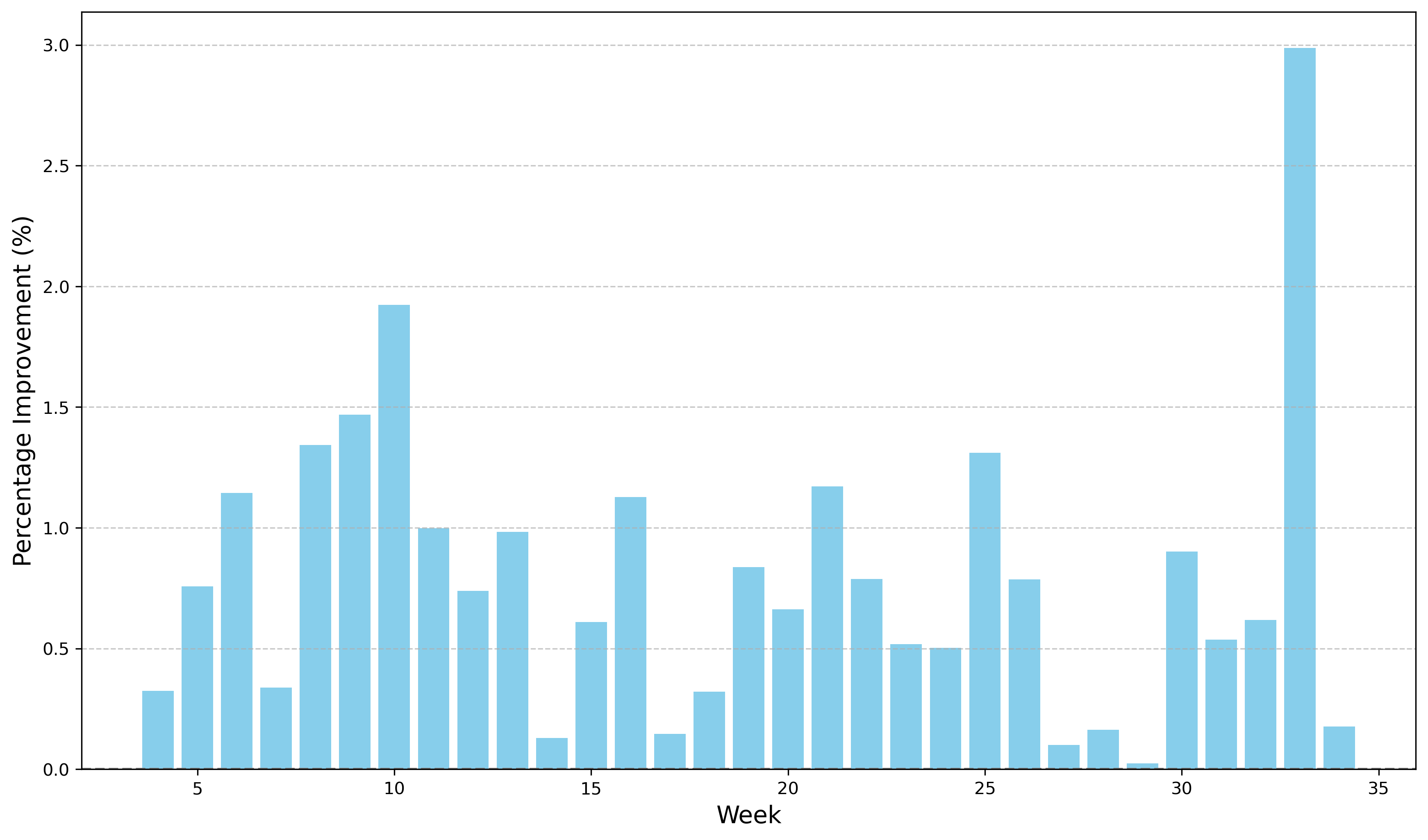}
    \caption{{\method} provides improvements over the baseline throughout the season.}
    \label{fig:improv-week}
\end{figure}

\section{Experimental Results}
\label{sec:results}

In this section we will present our data, the experimental setup and our results in detail.  

{\bf Data:} For our experiments we will make use of possession-level data. 
Our data cover the NBA seasons of 2022-23 and 2023-24 (both regular seasons and playoffs). 
Each data point corresponds to a single possession and includes the following tuple: ${\tt <p_i, offPl_{i,1}, offPl_{i,2}, offPl_{i,3}, offPl_{i,4}, offPl_{i,5}, defPl_{i,1}, defPl_{i,2}, defPl_{i,3}, defPl_{i,4}, defPl_{i,5}>}$, where ${\tt p_i}$ are the points scored in possession $i$, ${\tt offPl_{i,j}}$ is the $j^{th}$ player on offense during possession $i$ and ${\tt defPl_{i,j}}$ is the $j^{th}$ player on defense during possession $i$. 
We use the first season (2022-23) to obtain the players' RAPM ratings and the second season (2023-24) to evaluate the predictive power of {\method}. 
One of the issues we face with this set up is that there are players (e.g., rookies, that is, players that just entered the league from college or from other professional leagues) that appear in the 2023-24 season in our data but not in the season prior. 
In this case, we set their offensive and defensive RAPM to -1 each. 
In Section \ref{sec:conclusions} we will discuss some other possible approaches to addressing this challenge.

\begin{figure}[ht]
    \centering
    \includegraphics[width=0.75\linewidth]{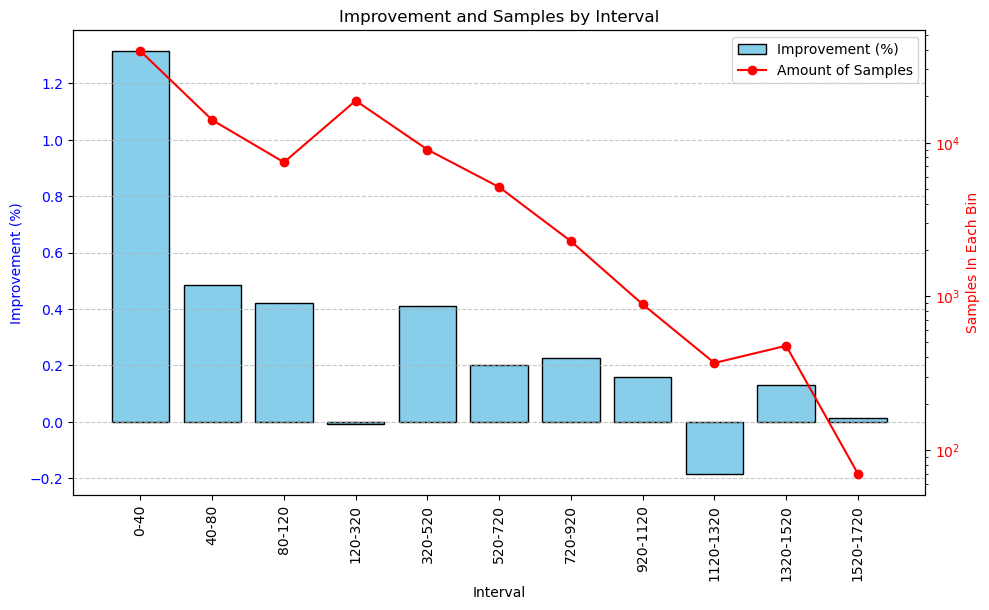}
    \caption{{\method}'s improvements over the baseline are larger when we have observed fewer data for the lineups making predictions for. The prior incorporated in our method allows {\method} to obtain lineup ratings with good predictive power with fewer data. }
    \label{fig:improv-sample}
\end{figure}

{\bf Experimental evaluation: }
Our objective is to evaluate the benefits of {\method} over the baseline usage of raw lineup ratings, in predicting the outcome of future matchups/possessions. 
As aforementioned the 2022-23 data will be used to obtain the RAPM player ratings that will be used for our lineup priors. 
Then we will use the 2023-24 data to evaluate the predictive ability of {\method} in an expanding window fashion. 
More specifically we will start with the first 4 weeks of the season and use these as our observations in order to make (out of sample) predictions for the possessions in week 5. 
From then on, we will keep expanding the window weekly and use weeks 1 through ($n$-1), to make predictions for the possessions observed in week $n$. 
This set up ensures that we are always performing out of sample evaluation, while also allowing us to examine how the prediction power of {\method} improves as we observe more data for a lineup. 
Our evaluation metric is the Root Mean Square Error (RMSE), and we calculate the RMSE for each week. 
Once we calculate the average RMSE for {\method} and the baseline, we then calculate the relative improvement $\improv_{\method}$ from {\method} as:

\begin{equation}
    \improv_{\method} = \dfrac{RMSE_{\method}-RMSE_{baseline}}{RMSE_{baseline}}
    \label{eq:improv}
\end{equation}

Figure \ref{fig:improv-week} depicts the improvement $\improv_{\method}$ for each of the weeks in our test set. 
As we can see, even though overall {\method} provides clear improvement over the baseline, there is some variability of its performance over the course of the season. 
However, there is not any specific trend observed with regards to the magnitude of the improvement from {\method} as the season progresses. 
The reason for this is that even though early in the season  every lineup has played very little time (since it is still early in the season) during each week later in the season there are lineups that might have played a few dozens of possessions, as well as, lineups that might have played several hundreds of possessions.  
To better understand the benefits of {\method} we calculate the improvement $\improv_{\method}$ as a function of the number of samples we have in our training set for {\method} at the time we make a prediction. 
This allows us to differentiate between a prediction during the last week of the season for a lineup that has only played 10 possessions during the season and one that has played 500 possessions. 
Figure \ref{fig:improv-sample} presents the results. 
As we can observe {\method} provides a larger benefit when making predictions for lineups for which we have fewer data points in the training data. 
These are also the majority of the cases in a real setting, as we mentioned in section \ref{sec:intro} and we can see from the sample sizes in each bin in figure \ref{fig:improv-sample}. 
For lineups where we have more than 500 possessions for instance, there is a high probability that this sample includes a good mix of opponents (in terms of strength) and thus, the raw rating has {\em converged} to - or close to - its true value. 
For lineups with little training data {\method}'s prior rating helps the model make a more informed prediction as compared to the noisy raw ratings.

\begin{figure}[ht]
    \centering
    \includegraphics[width=0.75\linewidth]{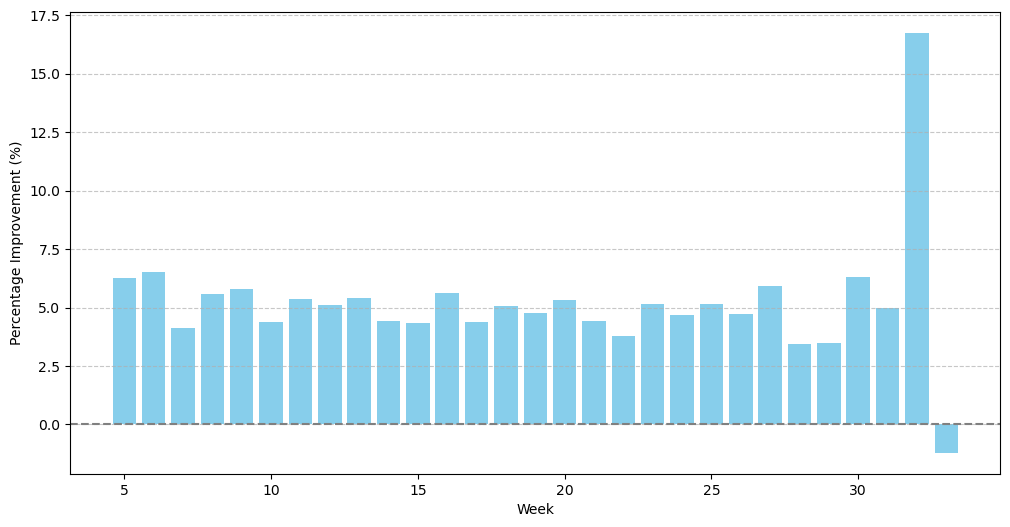}
    \caption{{\method} provides even higher improvements in terms of predictive power when predicting the points scored in possessions with previously unseen lineups.}
    \label{fig:improv-unseen}
\end{figure}

In fact, to further explore the predictive power of {\method}'s prior we calculate the RMSE for possessions that include lineups that we have not seen before in our training set (and which were excluded from the results presented above). 
This situations appear fairly frequently either because coaches want to experiment with new lineups, or due to injuries, or due to acquisition of new players during the season. 
Even though {\method} will not have a final regression rating for these lineups, since it has not appear during training, we can use the prior belief the model has on this lineup to make our predictions. 
Furthermore, there is no raw rating for these lineups since they have not played before. 
Therefore, the baseline method would use the league average points per possession for the prediction. 
Figure \ref{fig:improv-unseen} depicts the results and as we can see, {\method} consistently outpeforms the baseline, with a $\improv_{\method} \sim 5\%$. 
During the last 2 weeks we observe some outliers in terms of the performance, but it is worth noting that these weeks correspond to the playoffs and include very few games, and very few newly seen lineups.

These improvements might seem small, but even a 1.5\% improvement adds up quickly over the possessions of a whole game to approximately 3.4 points, which is not small. 
For comparison, this is higher than the home edge that betting markets incorporate in their handicap, which is about 2 points at the moment \cite{lopez2018often}. 
In the following section we will discuss various improvements that can push this gain even higher, but it should be evident that {\method} utilizes the player rating information (prior) in a beneficial way in terms of out-of-sample predictions. 

\section{Discussion and Conclusions}
\label{sec:conclusions}

In this work we design {\method}, a regression-based model to evaluate basketball lineups. 
{\method} addresses the major problem when rating lineups, that is, small sample size. 
A typical lineup will play an average of 30 to 40 total possession for a whole season and its observed performance is susceptible to variance. 
Our experiments indicate that {\method} provides benefits over current baselines, and these benefits are higher when the sample size of observed lineup performance is smaller (i.e., in those settings that we are most interested in). 

Even though our main objective for this work is to show that adjusting for opposition (through the model covariates) and for the players in the lineup (through the prior), one could further tweak the model to improve its predictive performance even more. 
In particular, currently we use player ratings obtained from the previous season. 
These ratings could also be continually updated as the current season progresses. 
While this will increase the computation time of {\method}, it will be more accurately capturing the performance of players at the moment. 
At the least, we will be able to obtain more accurate ratings (instead of setting them to -1) for players that were not in the league the previous season (either because they just entered the league, or because they were playing overseas etc.). 
Furthermore, we use a single regularization constant for both the offensive and defensive lineups. 
However, and similar to the RAPM player ratings, a separate regularization constant could further improve the predictive power of our method, since the offensive performance of a lineup/team tends to stabilize faster than the defensive one \cite{partnow2021midrange}.

\bibliography{collas2025_conference}

@book{partnow2021midrange,
  title={The midrange theory},
  author={Partnow, Seth},
  year={2021},
  publisher={Triumph Books}
}

@article{lopez2018often,
  title={How often does the best team win? A unified approach to understanding randomness in North American sport},
  author={Lopez, Michael J and Matthews, Gregory J and Baumer, Benjamin S},
  journal={The Annals of Applied Statistics},
  volume={12},
  number={4},
  pages={2483--2516},
  year={2018},
  publisher={JSTOR}
}

@misc{nba-eff,
  title = {NBA Advanced Stats: Lineup Efficiency},
  howpublished = {\url{https://stats.nba.com/lineups/advanced/}},
  note = {Accessed: 2025-02-18}
}

@inproceedings{pelechrinis2018linnet,
  title={Linnet: Probabilistic lineup evaluation through network embedding},
  author={Pelechrinis, Konstantinos},
  booktitle={Joint European Conference on Machine Learning and Knowledge Discovery in Databases},
  pages={20--36},
  year={2018},
  organization={Springer}
}

@article{deshpande2016estimating,
  title={Estimating an NBA player’s impact on his team’s chances of winning},
  author={Deshpande, Sameer K and Jensen, Shane T},
  journal={Journal of Quantitative Analysis in Sports},
  volume={12},
  number={2},
  pages={51--72},
  year={2016},
  publisher={De Gruyter}
}

@inproceedings{sill2010improved,
  title={Improved NBA adjusted+/-using regularization and out-of-sample testing},
  author={Sill, Joseph},
  booktitle={Proceedings of the 2010 MIT Sloan sports analytics conference},
  year={2010}
}

@book{winston2022mathletics,
  title={Mathletics: How gamblers, managers, and fans use mathematics in sports},
  author={Winston, Wayne L and Nestler, Scott and Pelechrinis, Konstantinos},
  year={2022},
  publisher={Princeton University Press}
}

@article{adjpm,
  title={Measuring How NBA Players Help Their Teams Win},
  author={Rosenbaum, Dan},
  journal={Available at: \url{http://www.82games.com/comm30.htm}. (Last accessed: 5-6-2018)},
  year={2004}
}
\bibliographystyle{collas2025_conference}


\end{document}